\documentclass[10pt,twocolumn,letterpaper]{article}

\usepackage{cvpr}      

\usepackage{xspace}
\usepackage{amssymb}
\usepackage{amsmath}
\usepackage{amsfonts}
\usepackage{wrapfig}
\usepackage{algorithm}
\usepackage{algorithmic}

\usepackage{multicol}
\usepackage{multirow}
\usepackage{makecell}
\usepackage{wrapfig}
\usepackage{tabularx}
\usepackage{adjustbox}

\usepackage{pifont}
\usepackage{color}
\usepackage{colortbl}

\definecolor{lightgray}{rgb}{0.8, 0.8, 0.8}
\definecolor{lgray}{rgb}{0.66, 0.66, 0.66}

\definecolor{lblu_tab}{RGB}{225, 235, 246}
\definecolor{orange_vitad}{RGB}{222, 131, 68}
\definecolor{blue_vitad}{RGB}{106, 153, 208}
\definecolor{trajectory_green}{RGB}{126, 171, 85}
\definecolor{trajectory_yellow}{RGB}{245, 194, 66}
\definecolor{tab_others}{RGB}{235, 235, 235}
\definecolor{tab_ours}{RGB}{225, 235, 246}

\definecolor{whit_tab}{RGB}{255, 255, 255}
\definecolor{gray_tab}{RGB}{246, 246, 246}
\definecolor{oran_tab}{RGB}{252, 242, 237}
\definecolor{blue_tab}{RGB}{227, 240, 251}

\definecolor{cvprblue}{rgb}{0.21,0.49,0.74}
\usepackage[pagebackref,breaklinks,colorlinks,allcolors=cvprblue]{hyperref}

\def\paperID{1762} 
\def\confName{CVPR}
\def\confYear{2025}

\title{SVP: Style-Enhanced Vivid Portrait Talking Head Diffusion Model}

\author{
    Weipeng Tan$^{*}$\textsuperscript{\rm 1}, Chuming Lin$^{*}$\textsuperscript{\rm 2}, Chengming Xu\textsuperscript{\rm 2}, Xiaozhong Ji\textsuperscript{\rm 2}, Junwei Zhu\textsuperscript{\rm 2},
    \\ Chengjie Wang\textsuperscript{\rm 2}, Yunsheng Wu\textsuperscript{\rm 2},  Yanwei Fu\textsuperscript{\rm 1}
\\
\textsuperscript{1} Fudan University, China
\quad
\textsuperscript{2} Youtu Lab, Tencent
}

\begin{document}

\twocolumn[{%
    \renewcommand\twocolumn[1][]{#1}%
    \maketitle
    \begin{center}
	\centering
	\captionsetup{type=figure}
	\includegraphics[width=1\linewidth]{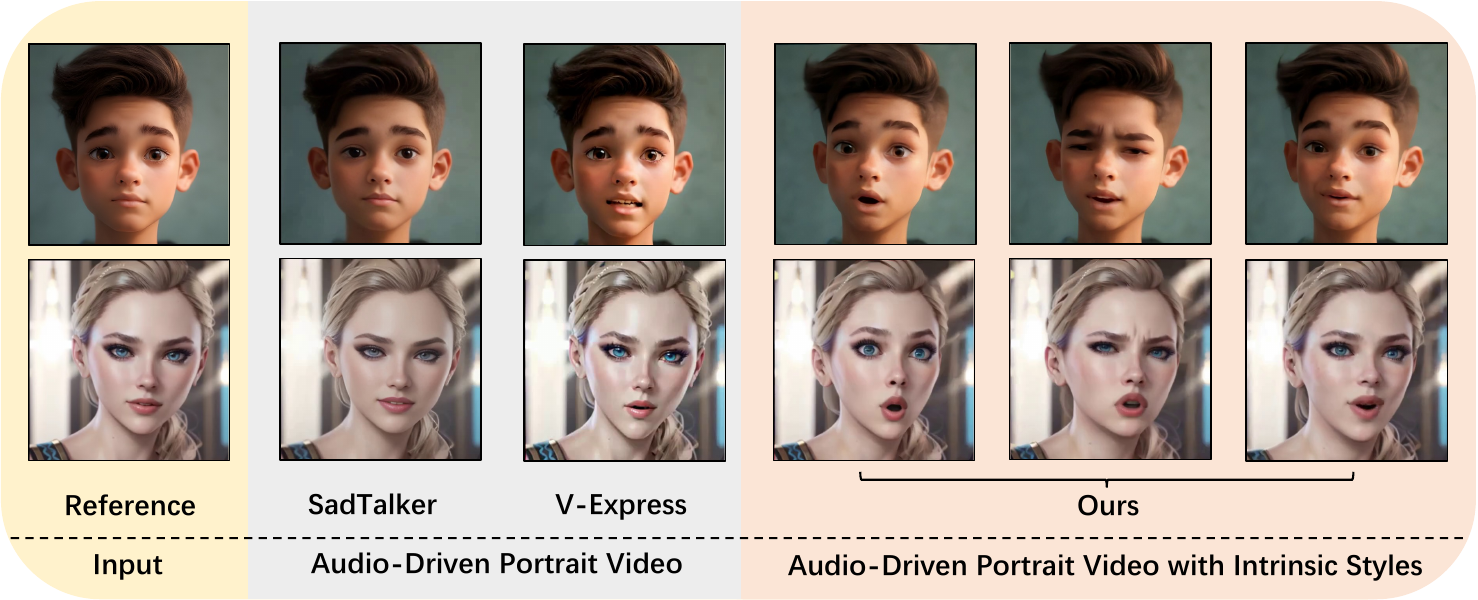} 
	\captionof{figure}{In talking head generation, given a audio and the reference image, both the GAN-based method SadTalker and the diffusion-based method V-Express have generated monotonous portrait videos, in which the primary movement is observed in the lips. In contrast, our approach is capable of generating diverse and vivid portrait videos based on varying intrinsic styles.}
	\label{fig:intro}
    \end{center}%
}]

\renewcommand{\thefootnote}{\fnsymbol{footnote}}
\setcounter{footnote}{0} 
\footnotetext[1]{Equal contributions. This work was done when Weipeng Tan was an intern at Tencent Youtu Lab.\\
Project page: \url{https://svportrait.github.io/}}
\newcommand{\name}{$\mathtt{SVP}$}

\begin{abstract}
Talking Head Generation (THG), typically driven by audio, is an important and challenging task with broad application prospects in various fields such as digital humans, film production, and virtual reality. While diffusion model-based THG methods present high quality and stable content generation, they often overlook the \textit{intrinsic style} which encompasses personalized features such as speaking habits and facial expressions of a video. As consequence, the generated video content lacks diversity and vividness, thus being limited in real life scenarios. To address these issues, we propose a novel framework named Style-Enhanced Vivid Portrait (\name) which fully leverages style-related information in THG. Specifically, we first introduce the novel probabilistic style prior learning to model the intrinsic style as a Gaussian distribution using facial expressions and audio embedding. The distribution is learned through the `bespoked' contrastive objective,  effectively capturing the dynamic style information in each video. Then we finetune a pretrained Stable Diffusion (SD) model  to inject the learned intrinsic style as a controlling signal via cross attention. 
Experiments show that our model generates diverse, vivid, and high-quality videos with flexible control over intrinsic styles, outperforming existing state-of-the-art methods.

\end{abstract}

\section{Introduction}
\label{sec:intro}

Recent advancements in generative models have shed light on generating high-quality and realistic videos under various controlling conditions such as texts~\cite{rombach2022ldm}, images~\cite{blattmann2023svd}, videos~\cite{wu2023tune}, etc. Among all the different subtasks of video generation, Talking Head Generation (THG), as a human-centric task which aims to generate videos of talking heads guided by conditions such as speech and images, has emerged as a significant problem due to its wide application in scenarios such as digital humans, film production, virtual reality. In spite of its importance, this is one of the most challenging tasks in video generation, resulted from its low tolerance to artifacts in general and its demand of high fidelity in lip shapes, facial expressions, and head motions. 



Following the commonly used generative models, the GAN-based THG methods~\cite{prajwal2020wav2lip, zhang2023sadtalker} have achieved remarkable results in generating high-resolution videos through adversarial training between generators and discriminators, particularly in terms of visual quality and lip-sync accuracy. 
Diffusion model-based THG methods~\cite{shen2023difftalk,tian2024emo}, on the other hand, excel in generating high-quality and high-resolution images and videos, and it outperforms GANs in terms of the stability and consistency of the generated content, thus becoming the mainstream methods for THG. 
These methods largely facilitate THG by strengthening the explicit controlling conditions such as facial keypoints and head motion sequences. However, they generally ignore the important fact of talking head videos. Essentially, when different people present speeches in real-life cases, they could have significant differences in habits and emotions under various circumstances. Such a fact in turn leads to different attributes in the corresponding talking head videos, including the visemes and expressions. Consequently, these habits and emotions are embedded as the \textit{intrinsic style} in talking head videos. This intrinsic style, while being highly related to whether a video is realistic, can hardly be inferred from conditions such as facial keypoints which are widely adopted by the previous methods. As a result, when there is a large gap in the intrinsic style between the reference face and the speaker of the style reference video, the previous methods struggle to reproduce the real situation accurately. 

To this end, we propose a novel framework named \name, which can effectively extract intrinsic style features with the assistance of audio information through a self-supervised method, and apply them to the generation of talking head videos in a manner suitable for diffusion models. This approach not only improves the overall quality of the generated videos, ensuring better synchronization and control but also accurately transfers facial expressions and individualized details to new faces.

Specifically, our \name~focuses on two main problems, i.e. extracting intrinsic style embeddings from style reference videos and controlling diffusion models with such embeddings. For intrinsic style extraction, a naive solution would be following StyleTalk~\cite{ma2023styletalk}, which maps 3D Morphable Model (3DMM)~\cite{blanz1999morphable} expression coefficients of the style reference video to style-related features. However, since attributes like visemes and expressions vary along the video frames, the deterministic embedding would suffer from insufficient capacity to model the latent manifold of intrinsic styles. Moreover, as one of the main parts of the video, the use of corresponding audio, which contains abundant information regarding the intrinsic styles, was not explored in StyleTalk, leading to unrepresentative style embeddings.

To solve these problems, we propose the novel Probabilistic Style Prior Learning as an alternative based on the transformer backbone. Concretely, the audio and visual information of each video interacts with each other in the transformer style encoder, which models the intrinsic style of this video as a Gaussian distribution with predicted mean and standard deviation. Through contrastive learning, the extracted features exhibit significant clustering across different identities and emotions, not only helping the model better understand the video content but also providing an effective way to capture and express the intrinsic style of individuals. After achieving the intrinsic style, it is integrated into the denoising process of target videos via additional cross attention, along with other conditions including the simplified facial keypoints for head movements and audio for lip shapes and movements around the mouth. Thanks to the design of the probabilistic style prior, we can resample from the predicted distributions to provide enough variation for the style-related information, thus resulting in the strong generalization ability of the trained model.





To validate the effectiveness of our proposed method, we conduct extensive experiments and comparisons on the MEAD~\cite{wang2020mead} and HDTF~\cite{zhang2021hdtf} datasets. Our method significantly outperforms other competitors across multiple metrics, including FVD~\cite{unterthiner2018fvd}, FID~\cite{heusel2017fid}, PSNR, SSIM, the offset and confidence of SyncNet~\cite{chung2017syncnet} and StyleSim. In addition to quantitative evaluation, we also perform comprehensive qualitative assessments. The results indicate that our method can generate highly natural and expressive talking videos, and can produce different emotions or even multiple changes in expressions within the same video according to user needs, achieving satisfactory visual effects.

Overall, our contributions are summarized as follows:
\begin{itemize}

\item We are the first to propose an audio-driven talking head generation framework based on a diffusion model that considers intrinsic style. \name~can generate realistic talking head videos with different intrinsic styles from the style reference videos. 

\item We propose an intrinsic style extractor that captures and expresses the intrinsic style of individuals via a self-supervised approach. 
We also incorporate audio information as an auxiliary into the style extractor to enhance the intrinsic style features. This allows the model to reflect the emotions and habits of speakers more accurately.

\item We designed a probabilistic style prior learning to adapt diffusion models. During the training of the style layer in the diffusion model, we sample the intrinsic style prior from the learned Gaussian distribution, enhancing stability and generalization capability.

\end{itemize}

\section{Related Work}
\label{sec:related_works}

\paragraph{GAN-Based Talking Head Generation.}
There has been significant research on GAN-based methods for person-generic audio-driven talking head generation. Early methods~\cite{prajwal2020wav2lip,cheng2022videoretalking,wang2023lipread} achieved lip synchronization by establishing a discriminator that correlates audio with lip movements. Other approaches~\cite{zhou2020makeittalk,zhou2021pcavs,wang2021audio2head,wang2022one,zhang2023sadtalker} generated portrait videos by mapping audio to key facial information, such as landmarks, key points, or 3D Morphable Model (3DMM)~\cite{blanz1999morphable} coefficients, before rendering the final frame. However, due to the limitations of GANs in terms of generative capacity, the results produced by these methods often suffer from artifacts like pseudo-textures or restricted motion ranges.

\paragraph{Diffusion Model-Based Talking Head Generation.}
Recently, there has been a surge of research ~\cite{shen2023difftalk,xie2024xportrait,tian2024emo,wang2024vexpress,wei2024aniportrait,xu2024hallo,yang2024megactor} utilizing diffusion models to achieve high-quality portrait videos. Among these, X-Portrait~\cite{xie2024xportrait} and MegActor~\cite{yang2024megactor} rely on the pose and expression from the source video to generate the target video, which limits their ability to produce videos based solely on audio. DiffTalk~\cite{shen2023difftalk} was the first to modify lip movements using audio and diffusion models, but it does not extend to driving other head parts. EMO~\cite{tian2024emo} was the first to leverage LDM~\cite{rombach2022ldm} and audio features to achieve overall motion in portraits. V-Express~\cite{wang2024vexpress} controls the overall motion amplitude by adjusting audio attention weights, while Hallo~\cite{xu2024hallo} designed a hierarchical module to regulate the motion amplitude of different regions. In summary, current audio-driven diffusion model approaches have not taken into account that each portrait should exhibit a corresponding style while speaking, which is essential for generating higher-quality portrait videos.

\paragraph{Stylized Talking Head Generation.}
Previous research has explored several GAN-based methods ~\cite{wang2020mead,wu2021imitating,ji2021evp,ji2022eamm,liang2022expressive,sinha2022emotion,ma2023styletalk} for extracting style information to apply in talking head generation. MEAD~\cite{wang2020mead} and Emotion~\cite{sinha2022emotion} directly inject style labels into the network to drive the corresponding emotions. GC-AVT~\cite{liang2022expressive} and EAMM~\cite{ji2022eamm} map the facial expressions of each frame in the source video to each frame in the target video. LSF~\cite{wu2021imitating} and StyleTalk~\cite{ma2023styletalk} employ 3D Morphable Models (3DMM) to extract facial information and construct style codes that drive the desired styles. Building on these approaches, our framework is the first to propose the extraction of intrinsic style priors by integrating audio and 3D facial information. Additionally, we are first to design a probabilistic learning to enhance style control within diffusion models.
\section{Method}
\begin{figure*}[t!]
	\centering
	\includegraphics[width=2.1\columnwidth]{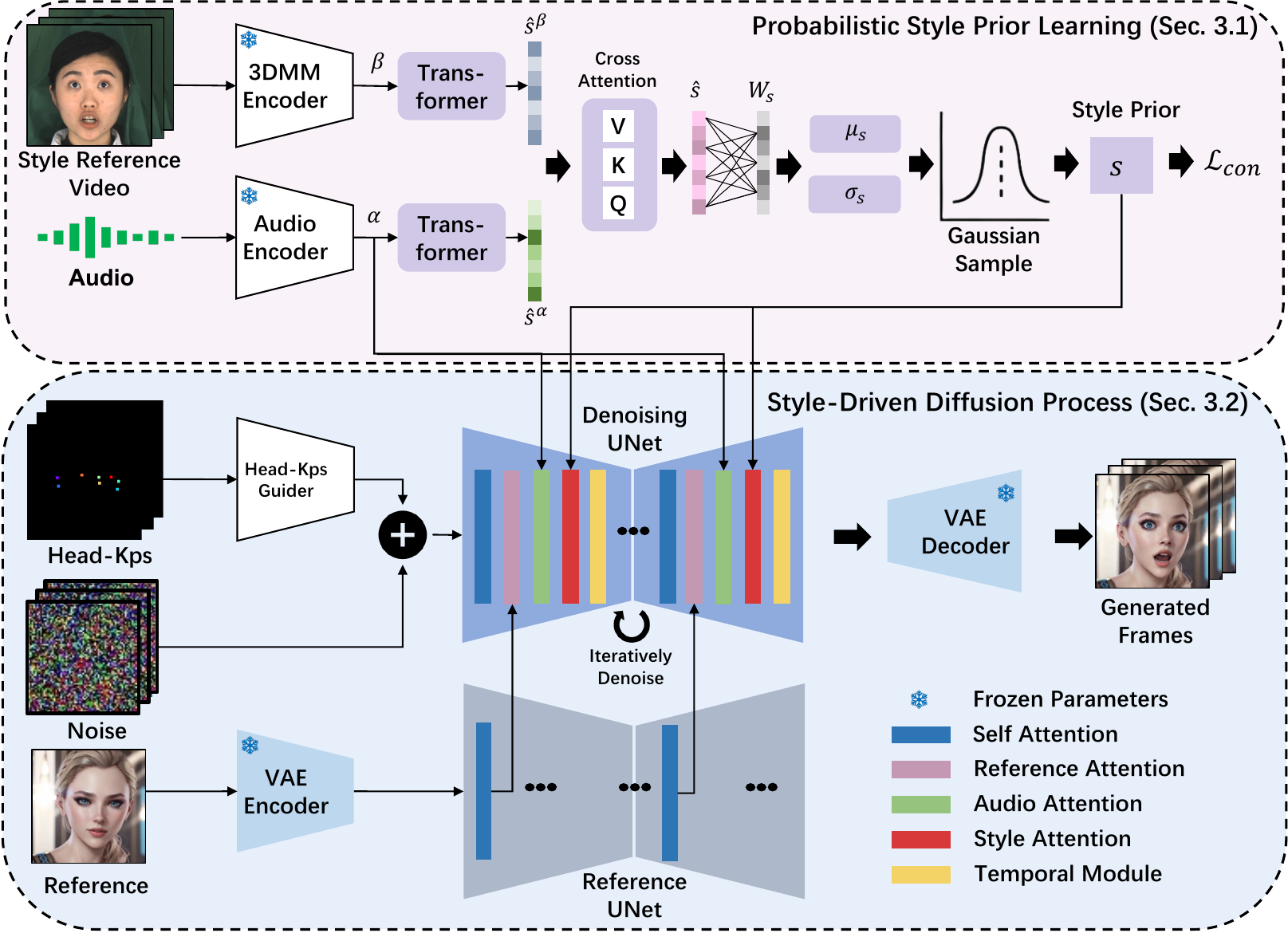} 
	\caption{\textbf{The Framework of \name.} Our \name~Framework includes Probabilistic Style Prior Learning and Style-Driven Diffusion Process. In Probabilistic Style Prior Learning, we utilizes a dual-branch transformer to convert the audio features $\alpha$ and the expression parameters $\beta$ into the latent vectors $\hat{s}^{\alpha}$ and $\hat{s}^{\beta}$ respectively, then obtained the style-related embedding $\hat{s}$ via the cross attention layer. Finally, we uses the learnable parameter $W_s$ to map the embedding $\hat{s}$ to mean $\mu_s$ and variance $\sigma_s$, and the style prior $s$ is sampled by $\mathcal{N}(\mu_s, \sigma_s^2)$. In Style-Driven Diffusion Process, the Denoising UNet takes the reference image, Head-Kps sequences, audio features and style prior as conditions to denoise the input noise at each time step.
	} 
	\label{fig:framework}

\end{figure*}

\paragraph{Problem Formulation.}
The goal of THG is to generate a talking head video under the control of a reference portrait image, audio, Head-Kps image sequence, and intrinsic style prior. Among these conditions, the reference portrait image provides the background and facial identity, the audio guides the lip movements, and each Head-Kps image controls the head position and pose for each generated frame. The Head-Kps images are synthesized by mapping 8 facial keypoints onto a black background. These 8 facial keypoints correspond to the left and right edges of the face, the pupils of both eyes, and the bridge of the nose, which are used to guide the overall head movement. In addition, we propose probabilistic style prior learning to extract the style prior from the visual and audio content of a style reference video, which is used to determine facial emotion and speaking habit.


\paragraph{Preliminaries.}
In \name, we employ a Latent Diffusion Model (LDM)~\cite{rombach2022ldm} to generate video frames. The LDM uses a diffusion and denoising process in the latent space via a Variational Autoencoder (VAE). It maps the input image $x$ to the latent space, encoding the image as $z=E(x)$, which helps maintain visual quality while reducing computational cost. During the diffusion process, Gaussian noise $\epsilon \sim \mathcal{N}(\mathbf{0},\mathbf{I})$ is gradually introduced into the latent $z$, degrading it into complete noise $z\sim\mathcal{N}(\mathbf{0},\mathbf{I})$ after $T$ steps. In the reverse denoising process, the target latent $z$ is iteratively denoised from the sampled Gaussian noise using the diffusion model and then decoded by the VAE decoder $D$ into the output image $x=D(z)$. During training, given the latent $z_0 =E(x_0 )$ and condition $c$, the denoising loss is:
\begin{equation}
\mathcal{L}_{{denoising}}=\mathbb{E}_{\mathbf{z}_t,\boldsymbol{\epsilon},\mathbf{c},t}\|\boldsymbol{\epsilon}_\theta(\mathbf{z}_t,\mathbf{c},t)-\boldsymbol{\epsilon}_t\|^2.
\label{eq:denoise}
\end{equation}
Among them, $z_t = \sqrt{\alpha_t}z_0 + \sqrt{1-\alpha_t}\epsilon_t$  represents the noisy latent variables at timestep $t\in[1,T]$, and $\epsilon_t$  is the added noise. $\epsilon_\theta$ is the noise predicted by the UNet model, modified using an attention mechanism with parameters $\theta$. This model employs a cross-attention mechanism to fuse the condition $c$ with the latent features $z_t$ , thereby guiding the image generation.
\name~uses Stable Diffusion v1.5 (SDv1.5), a text-to-image Latent Diffusion Model (LDM), as the backbone. SDv1.5 is implemented based on U-Net~\cite{ronneberger2015unet}, with each Transformer~\cite{vaswani2017attention} block containing both self-attention and cross-attention layers. 

\paragraph{Overview.}
As depicted in Figure~\ref{fig:framework}, Our \name~consists of two important designs, namely Probabilistic Style Prior Learning and Style-Driven Diffusion Process. In Probabilistic Style Prior Learning, we propose the style extractor takes the 3DMM expression parameters and audio features as inputs to generate the style prior, which is represented as a Gaussian distribution. 
In the style-driven diffusion process, the encoded Head-Kps sequence, reference image, audio features, and style prior are progressively input into the Denoising UNet as control conditions through their respective attention layers. Finally, the vivid portrait frames are generated by the VAE decoder after the iteration of the denoising process.

\subsection{Probabilistic Style Prior Learning \label{sec:style}}
In order to learn representative intrinsic style indicators from style reference videos, we propose the novel probabilistic style prior learning. Built upon the transformer-based style encoder as in StyleTalk~\cite{ma2023styletalk}, we adopt a novel framework to make better usage of the style-related information contained in each video. Concretely, for a video clip, we first transform it into its corresponding frame-level audio parameters $\alpha\in\mathbb{R}^{N\times1920}$ via Whisper-Tiny~\cite{radford2023whisper} and sequential expression parameters $\beta\in{\mathbb{R}}^{N\times 64}$ via the 3DMM encoder, where $N$ denotes number of frames. These two modalities are then processed with a dual-branch transformer model as shown in Figure~\ref{fig:framework}, which outputs their counterparts $\hat{s}^{\alpha}, \hat{s}^{\beta}\in\mathbb{R}^{N\times d_s}$, where $d_s$ denotes feature channels. After achieving features for each modality, we interact with them with cross attention, leading to a style-related embedding $\hat{s}$ aware of both audio and visual information. 

With $\hat{s}$, we can then model the intrinsic style prior for each video as a Gaussian distribution. Specifically, an attention-based aggregation strategy is employed on $\hat{s}$ as follows:
\begin{align}
    \mu_s &= \mathrm{softmax}(W_s\hat{s})\cdot \hat{s}^T,\\
    \sigma_s^2 &= \mathrm{softmax}(W_s \hat{s})\cdot(\hat{s}^T-\mu_s)^2,\\
    s &= \mu_s + \sigma_s \cdot \epsilon, \quad \epsilon \sim \mathcal{N}(\mathbf{0}, \mathbf{I}),
\end{align}
where $W_s\in\mathbb{R}^{1\times d_s}$ is a trainable parameter, $\mu_s, \sigma_s^2$ denotes the mean and variance of the learned style prior $s$.

Compared with the naive style encoder used in StyleTalk~\cite{ma2023styletalk}, our proposed model mainly enjoys the following merits: (1) As mentioned in Sec.~1, the audio information is vital for extracting intrinsic style, while StyleTalk cannot handle such a modality. Moreover, audio typically contains primarily information about the spoken content, thus making it non-trivial to extract information that is complimentary to the visual information contained in video frames. In comparison with StyleTalk, we design a specific structure to handle these complex data, considering both visual and audio information, leading to stronger style embedding. (2) Since the emotion of speakers would change as the video frames go on, it is sufficient to represent the intrinsic style with a deterministic feature, i.e. the same way as in StyleTalk. Our method, on the other hand, learns a better sequential embedding, which helps us model the style prior as a Gaussian distribution that is more representative.



\subsection{Style-Driven Diffusion Process}
After learning intrinsic style from style reference videos, we can control details such as facial expressions with such a condition, enabling a more refined talking head generation process. Specifically, we build a talking head generation model based on previous methods such as V-Express~\cite{wang2024vexpress}. These methods apply various techniques to pretrained Stable Diffusion (SD) for better quality. For instance, ReferenceNet can generate similar feature maps and integrate the extracted features into the diffusion backbone, preserving the visual information of the face and background from the reference image. The audio projection module, which embeds audio information, controls the generation of lip movements through a cross-attention mechanism. The temporal attention layer, which enhances temporal coherence by performing self-attention on the frame sequence to capture inter-frame correlations. In addition to these existing methods, we further propose two novel modules named HEAD-Kps Guider and Style Projection as follows that can better facilitate the input data.

\paragraph{HEAD-Kps Guider.} 
Each HEAD-Kps image spatially corresponds to the respective target frame, containing information about the head's position and rotation. To fully utilize this, we use the HEAD-Kps Guider to encode the HEAD-Kps images. The Head-Kps images are constructed using landmarks from the upper half of the face, consisting of 8 keypoints. The HEAD-Kps Guider is a lightweight convolutional model that encodes the keypoints into HEAD-Kps features, which represent spatial information and match the shape of the latent features. Subsequently, before being input into the denoising U-Net, the multi-frame latent features are directly added to the corresponding encoded HEAD-Kps features, enabling the model to accurately interpret the head's spatial information.

\paragraph{Style Projection.} \label{sec:style-layer}
To utilize the intrinsic style priors obtained in Sec.~\ref{sec:style} to guide the denoising process, we first resample a corresponding intrinsic style prior $s$ from the Gaussian distribution learned from the style reference video. Then $s$ is injected into the diffusion UNet through an additional style attention layer, where it interacts with other features via a cross-attention mechanism to supplement additional facial details such as expressions and speaking habits.
The intrinsic style prior information can be injected into the spatial cross-attention layer to provide spatial knowledge as follows:
\begin{equation}
z_{s}= z_a + \mathrm{CrossAttn}\left(Q(z_a),K(s),V(s)\right),
\label{eq:style}
\end{equation}
where $z_a$ is the spatial latent features after being injected with reference attention and audio attention, and $z_s$ is the adjusted spatial features guided by intrinsic style prior spatial-aware level.

\subsection{Training Strategies}
\paragraph{Training of Intrinsic Style Extractor.} Essentially, codes with similar intrinsic styles should cluster together in the style space. Therefore we apply contrastive learning to the style priors by constructing positive pairs $(s, s^p)$ with the same identity and emotion, and negative pairs  $(s, s^n)$ with different identities or emotions. Then, the InfoNCE loss~\cite{chen2020simclr} with similarity metric $\zeta$ is enhanced between positive and negative sample pairs:
\begin{align}
    \omega(\tilde{s}) &= \exp(\zeta(s, \tilde{s}) / \tau), \\
    \mathcal{L}_{con} &= -\log\left(\frac{\omega(s^p)}{\omega(s^p)+\sum_{s^n\in\mathcal{S}^n}\omega(s^n)}\right),
\end{align}
where $\tau$ denotes a temperature parameter, $\mathcal{S}^n$ denotes all negative samples for $s$, and the similarity $\zeta(s_i, s_j)=\frac{1}{\|s_i-s_j\|_2+1}\in (0,1]$ is an improved version obtained as the inverse of the $\mathcal{L}_2$ distance between sample pairs. We additionally add a fixed constant to stabilize the numerical range of the similarity and make the training process more stable.

In the training of the intrinsic style extractor, we directly train all parameters of this lightweight model. Meanwhile, we use a random dropout trick when inputting the 3DMM expression coefficients $\beta$ and audio features $\alpha$ by setting some of the input 3DMM expression coefficients $\beta$ or audio features $\alpha$ to zero. This allows the model to obtain the style prior through a single modality.

\paragraph{Finetuning Diffusion Model.}
The training of the diffusion model adopts a three-stage progressive training method to gradually improve the model's generative capability and stability, with the noise prediction loss as in Eq.~\ref{eq:denoise} employed in each stage. (1) First, we train the model for single-frame image generation, where the diffusion UNet, ReferenceNet, and Head-Kps guider are involved in the training. In this stage, the diffusion UNet takes a single frame as input, the ReferenceNet processes different frames randomly selected from the same video clip, and the Head-Kps guider incorporates the encoded Head-Kps features into the latent space. Both the diffusion UNet and ReferenceNet initialize their weights from the original SD. (2) Second, we train the model for continuous multi-frame image generation, which includes the temporal module and the audio layer. In this stage, $f$ consecutive frames are sampled from a video clip and the parameters of ReferenceNet and Head-Kps guider are frozen. The temporal module initializes its weights from AnimateDiff~\cite{guo2023animatediff}. (3) After that the final stage is for transferring intrinsic style. In this stage, all other modules of the model are frozen, and only the style attention module is trained. This allows the model to generate corresponding facial expressions and details based on the intrinsic style input during the portrait image generation process.

\section{Experiments}

\begin{figure*}[t!]
    \centering
    \begin{subfigure}{1\columnwidth}
        \includegraphics[width=1.05\linewidth]{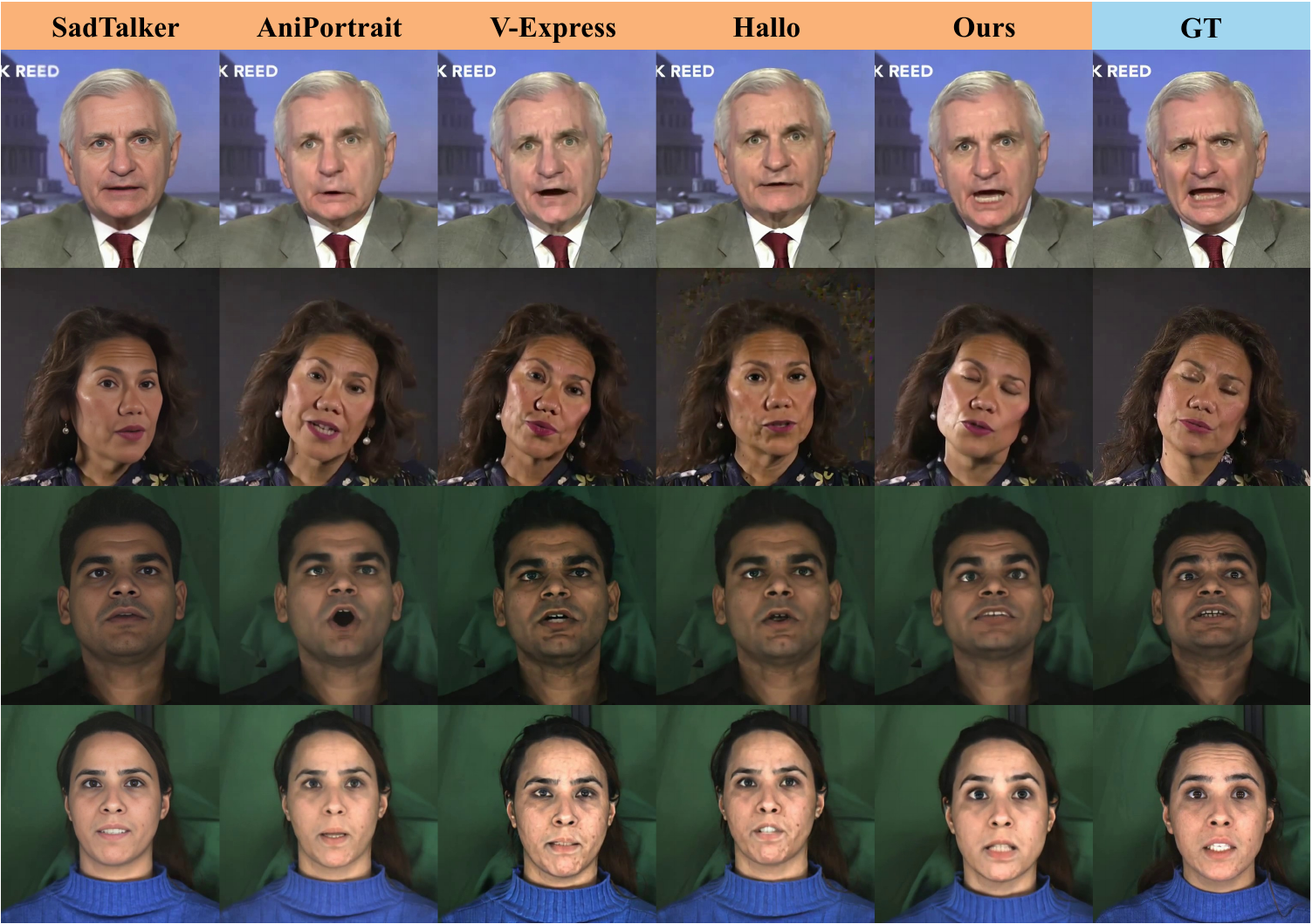} 
	\caption{}
	\label{fig:stoa_cmp}
    \end{subfigure}
    \hfill
    \begin{subfigure}{1\columnwidth}
        \includegraphics[width=1.05\linewidth]{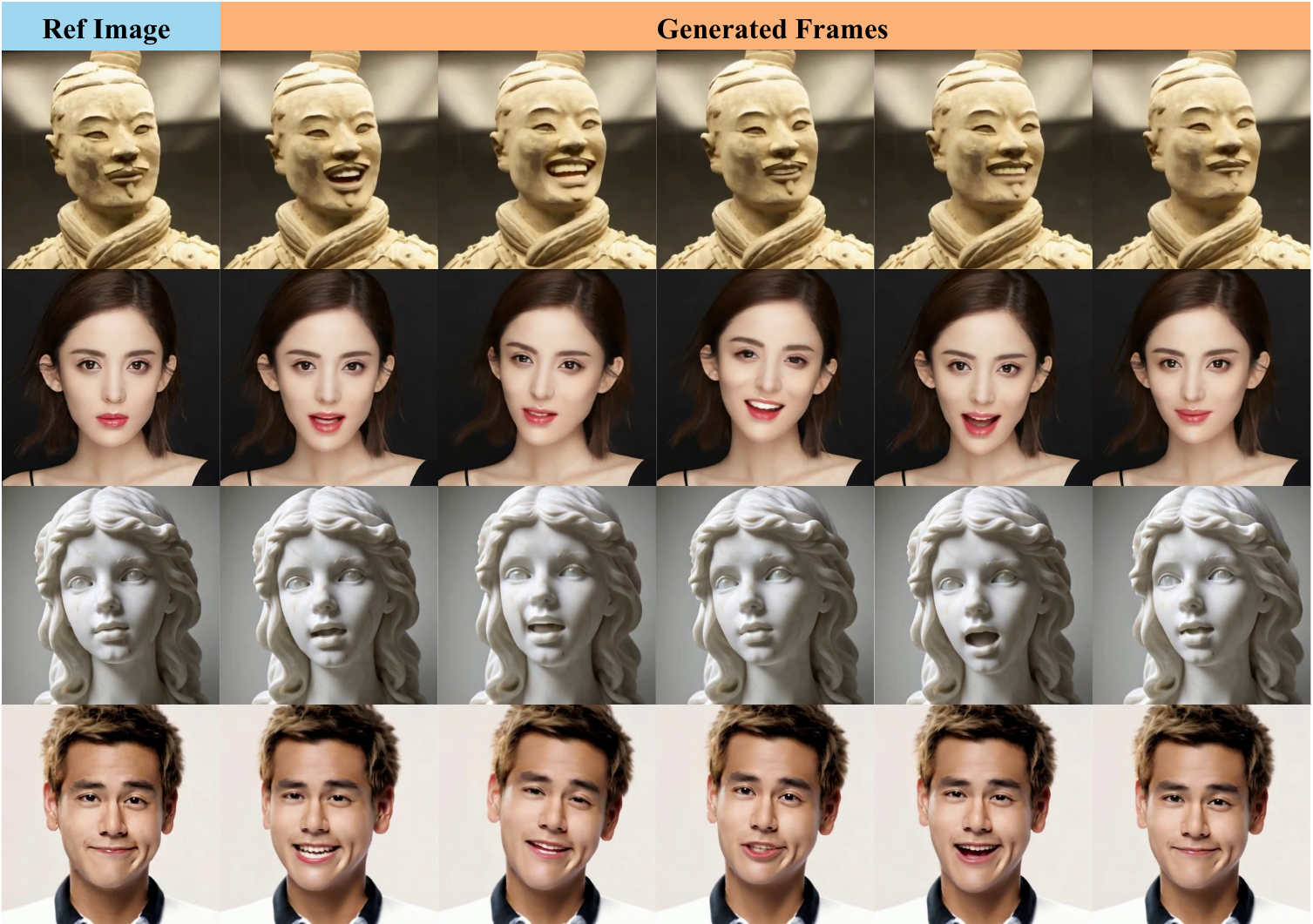}
        \caption{}
        \label{fig:demos}
    \end{subfigure}
    \caption{\textbf{(a)} Visual comparison with recent SOTA methods. The first two rows show the comparison of reconstruction results, while the last two rows show the comparison of intrinsic style transfer results. \textbf{(b)} Our method uses intrinsic style to generate frames on different types of portraits. This demonstrates that our method can successfully apply intrinsic style to various types of data, even if only real-life videos are available for training.}  
    \label{fig:two_images}
\end{figure*}

\begin{table*}[t!]
\footnotesize
\centering
\caption{The quantitative results of video reconstruction on the MEAD and HDTF dataset.}
\label{tab:reconstruction}
\setlength{\tabcolsep}{0.9mm}{
\begin{tabular}{ccccccccccccccc}
\toprule
            & \multicolumn{7}{c}{MEAD}                                                                            & \multicolumn{7}{c}{HDTF}                                                                              \\ \cmidrule(r){2-8}  \cmidrule(r){9-15}
Method      & FVD$\downarrow$            & FID$\downarrow$           & PSNR$\uparrow$          & SSIM$\uparrow$           & F-LMD$\downarrow$         & M-LMD$\downarrow$    & SyncNet$\downarrow$$\uparrow$     & FVD$\downarrow$            & FID$\downarrow$           & PSNR$\uparrow$          & SSIM$\uparrow$           & F-LMD$\downarrow$          & M-LMD$\downarrow$   & SyncNet$\downarrow$$\uparrow$       \\ \midrule
V-Express   & 340.01 & 33.66 & 27.51 & 0.8939 & 5.28 & 8.12  & 2.10/4.59 & 125.36 & 12.33 & 26.35 & 0.8688 & 30.58 & 35.56 & 2.70/6.05 \\
Hallo       & 445.49 & 35.06 & 25.84 & 0.8666 & 9.60 & 12.42 & 1.70/\textbf{5.94} & 231.96 & 15.31 & 22.44 & 0.8078 & 31.86 & 35.63 & 2.10/\textbf{6.74}\\
AniPortrait & 460.04 & 40.65 & 27.16 & 0.8982 & 5.86 & 9.03  & 8.30/1.40 & 292.18 & 13.27 & 25.27 & 0.8616 & 31.20 & 36.37 & 4.00/0.53\\ 
EAMM   & 529.34   & 58.77  & 21.88  & 0.8139  & 11.88  & 12.43   & 1.90/3.98 & 797.23 & 43.46 & 18.47 & 0.7202 &29.93 & 30.48  & 2.40/4.54\\
SadTalker   & 492.40  & 52.94 & 27.33 & 0.8863 & 7.23 & 11.03 & 0.40/4.74 & 536.68 & 23.7  & 22.62 & 0.8042 & 28.81 & 32.60 & 0.50/5.96\\  \midrule
Ours &
  \textbf{235.70} &
  \textbf{28.80} &
  \textbf{28.87} &
  \textbf{0.9126} &
  \textbf{4.64} &
  \textbf{6.45} &
  \textbf{0.30}/4.76 &
  \textbf{102.40} &
  \textbf{10.39} &
  \textbf{26.38} &
  \textbf{0.8800} &
  \textbf{22.95} &
  \textbf{26.01} & \textbf{0.20}/5.40 \\ \bottomrule 
\end{tabular}}
\end{table*}

\begin{table}[t!]
\small
\centering
\caption{The quantitative results of intrinsic style transfer on the MEAD dataset.}
\label{tab:tab-cross}
\resizebox{0.48\textwidth}{!}{
\setlength{\tabcolsep}{0.8mm}{
\begin{tabular}{@{}ccccccccc@{}}
\toprule
Method      & {FVD$\downarrow$}   & {FID$\downarrow$}  & {PSNR$\uparrow$} & {SSIM$\uparrow$}  & {F-LMD$\downarrow$} & M-LMD$\downarrow$ & SyncNet$\downarrow$$\uparrow$ & StyleSim$\uparrow$ \\ \midrule
V-Express   & \textbf{367.91}          & 63.17          & 21.23          & 0.7904          & 7.41           & 9.33  & 2.03/4.25 & 0.7819        \\
Hallo       & 399.18          & 59.05          & 19.70          & 0.7573          & 26.16          & 28.54  & 1.93/4.85  & 0.7851      \\
AniPortrait & 535.53          & 68.18          & 19.34          & 0.7583          & 26.55          & 29.81  & 8.67/1.32  & 0.7711      \\
EAMM   & 550.90 & 63.94 & 21.25 & 0.7848 & 12.51 & 12.72     & 2.17/4.20  & 0.7349     \\
SadTalker   & 492.83          & 82.71          & 20.07          & 0.7670          & 27.00          & 29.55    & \textbf{0.70}/\textbf{4.93} & 0.7417     \\
Ours w/o style & 485.47          & 57.96 & 21.73 & 0.8151 & 8.49  & 10.20 & 0.93/4.03 & 0.7879  \\ \midrule
Ours        & 392.31 & \textbf{56.69} & \textbf{22.07} & \textbf{0.8210} & \textbf{6.76}  & \textbf{8.44} & 0.77/4.14 & \textbf{0.8473} \\ \bottomrule
\end{tabular}}}

\end{table}


\begin{table}[t!]
\small
\centering
\caption{Comparison of intrinsic style clustering strength on different emotions. The higher value means the better clustering effects.}
\label{tab:cluster_emo}
\resizebox{0.45\textwidth}{!}{%
\begin{tabular}{@{}cccccc@{}}
\toprule
Input         & angry & contempt & disgusted & happy \\ \midrule
audio         & 2.44  & 2.12     & 2.51 & 2.49  \\
style         & 6.25  & 4.89     & 7.45 & 8.07  \\
style + audio & \textbf{6.57}  & \textbf{5.64}     & \textbf{7.63} & \textbf{8.38}  \\ \bottomrule
\end{tabular}}
\end{table}

\begin{figure}[t!]
	\centering
        \small
	\includegraphics[width=0.95\columnwidth]{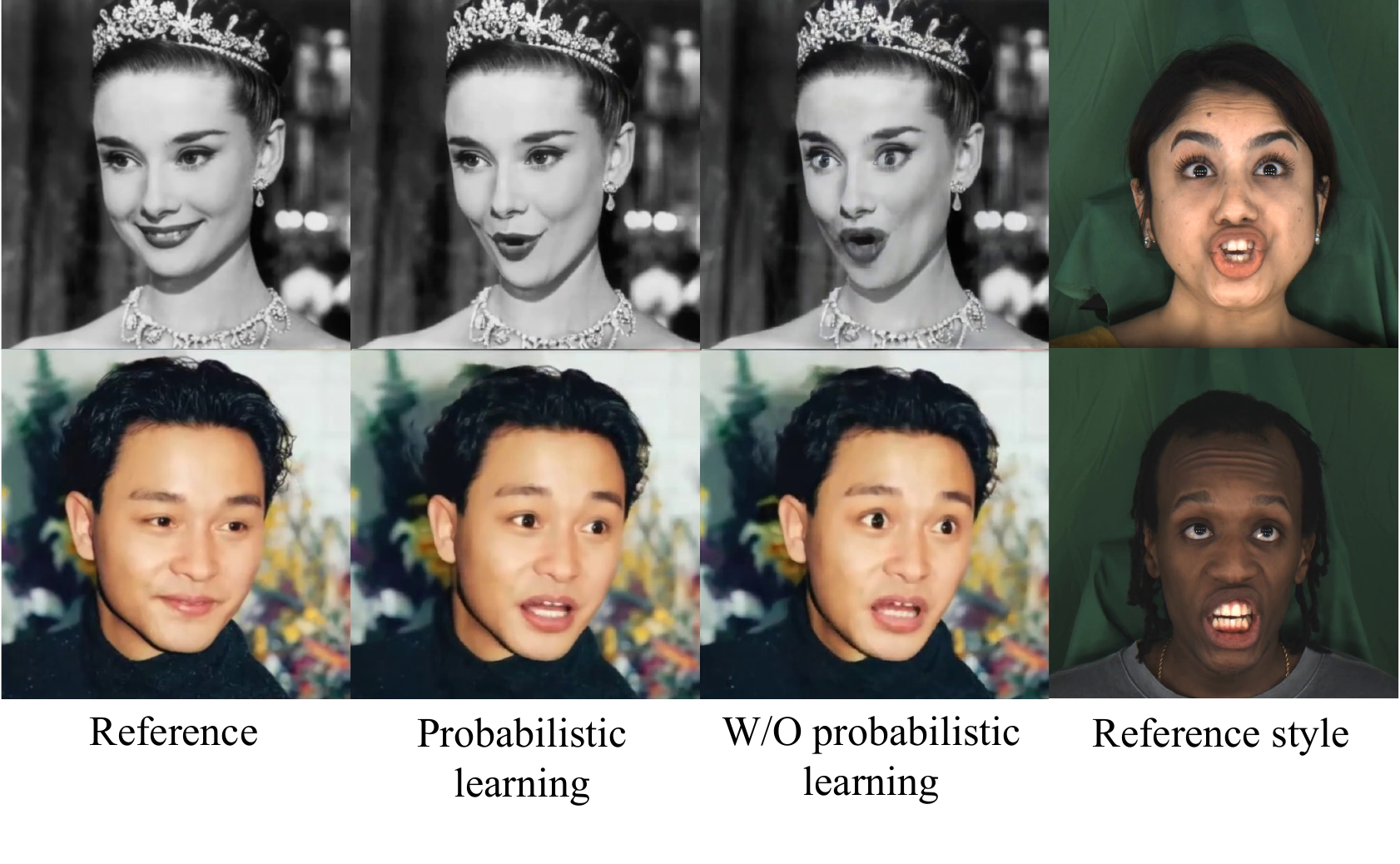} 
	\caption{Comparison of visualization results with and without Probabilistic Style Prior Learning.} 
	\label{fig:cmp}
\end{figure}

\begin{figure}[t!]
	\centering
	\includegraphics[width=1.0\columnwidth]{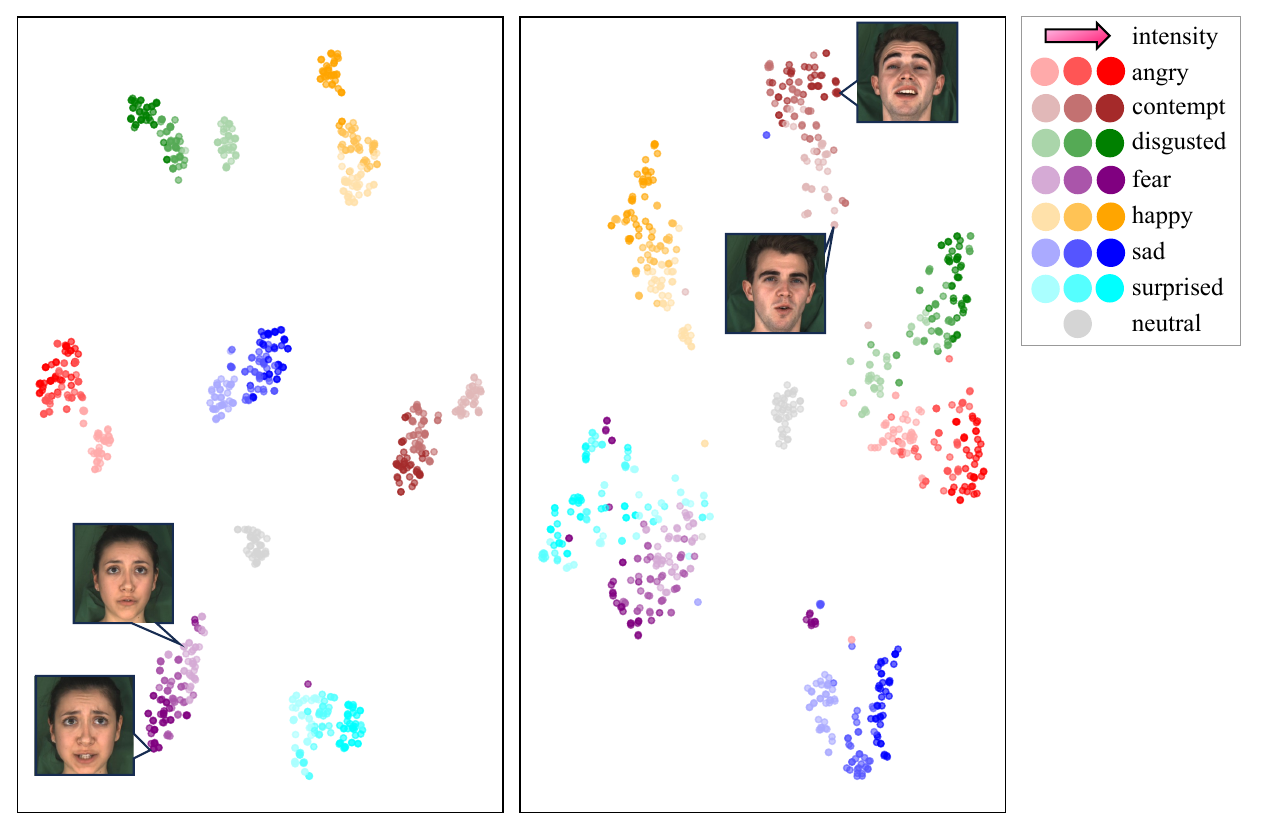} 
	\caption{Intrinsic style prior visualization. The color gets darker as the intensity of the emotion increases.} 
	\label{fig:cluster}
\end{figure}

\begin{figure}[t!]
	\centering
        \small
	\includegraphics[width=1.0\columnwidth]{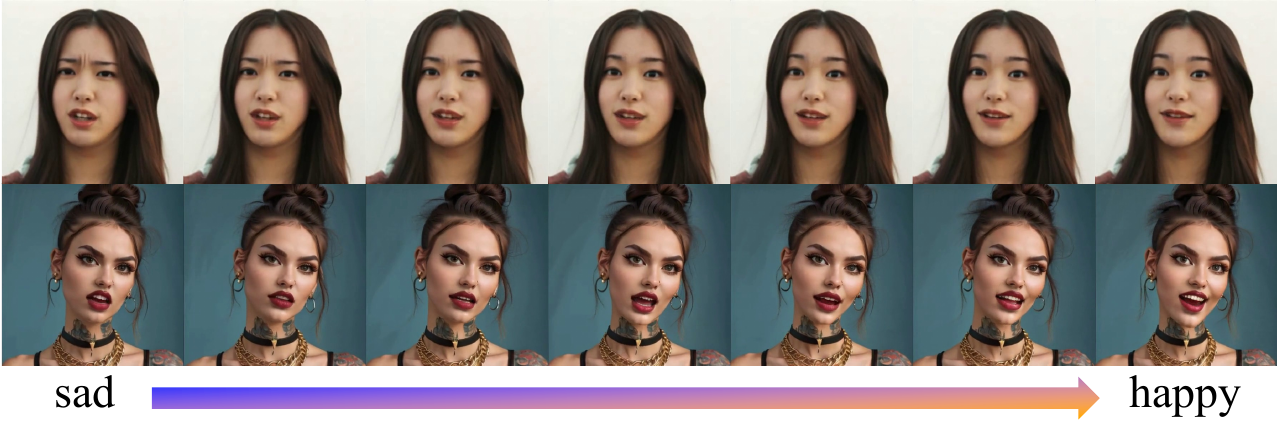} 
	\caption{Interpolation results between sad and happy emotions by controlling the intrinsic style prior.} 
	\label{fig:interpolation}
\end{figure}

\subsection{Experiments Setting}

\name~is implemented using PyTorch~\cite{paszke2019pytorch} and optimized with Adam~\cite{kingma2014adam}.
The intrinsic style encoder is trained on the MEAD~\cite{wang2020mead} and HDTF~\cite{zhang2021hdtf} datasets. 
During training, we consider samples with the same identity and emotion in MEAD as positive samples, and segments from the same video in HDTF as positive samples. Additionally, we will randomly dropout expression coefficients or audios, but they will not be zeroed out simultaneously.

The denoising UNet is trained on the MEAD, HDTF, and other videos from Internet. The facial regions in these videos are cropped and resized to 512×512. The total training dataset comprises approximately 300 hours of video.

In the multi-frame training stage, the number of consecutive frames $f$ is set to 8.
In the training of style projection, to enhance the generalization ability, we adopt different emotions for the same identity on the MEAD dataset (e.g. generating a sad video clip from a happy reference image). 

For training and testing set splitting, we select 10 identities out of 46 for testing on MEAD. As for HDTF, we randomly select 25 videos for testing. Precautions are taken to ensure that there is no overlap of character identities between the training and testing sets. During inference, to ensure fairness, we utilize EulerDiscreteScheduler as diffusion sampler with the denoising steps set as 25 for all diffusion-based methods.

\subsection{Quantitative Comparison}
We compare our method with several previous works, including EAMM~\cite{ji2022eamm}, SadTalker~\cite{zhang2023sadtalker}, AniPortrait~\cite{wei2024aniportrait}, V-Express~\cite{wang2024vexpress}, and Hallo~\cite{xu2024hallo}.

To demonstrate the superiority of the proposed method, we evaluate the model using several quantitative metrics. We utilize the Fréchet Inception Distance (FID)~\cite{heusel2017fid} to assess the quality of the generated frames, and further employe the Fréchet Video Distance (FVD)~\cite{unterthiner2018fvd} for video-level evaluation. To evaluate the quality of the generated talking head videos, Peak Signal-to-Noise Ratio (PSNR) and Structural Similarity Index Measure (SSIM) are adopted. To evaluate lip-sync accuracy, we use the Mouth Landmark Distance (M-LMD)~\cite{bulat2017far} and the average visual-audio offset and confidence of SyncNet~\cite{chung2017syncnet}. For assessing the accuracy of the generated facial expressions, we use the Full-Face Landmark Distance (F-LMD). 

Additionally, we introduce Intrinsic Style Similarity (StyleSim) to evaluate the performance of the generated results in terms of facial expressions and details. For the generated video $V_{res}$ and $V_{gt}$, we use the 3DMM encoder to extract their sequential expression parameters $\beta_{res}$ and $\beta_{gt} $. Then, we use the pretrained style encoder from StyleTalk~\cite{ma2023styletalk} to encode them into style representations $s_{res}$ and $s_{gt}$. We consider the cosine similarity between $s_{res}$ and $s_{gt}$ as StyleSim.

As shown in Table~\ref{tab:reconstruction}, in the video reconstruction experiments, our method achieves the best performance in most metrics on both MEAD and HDTF datasets. We have a significant advantage in evaluating the quality of video and single-frame images, as evidenced by the lower FVD and FID scores. The SSIM and PSNR scores indicate that the quality of the videos reconstructed by our method is significantly better than that of other methods. The M-LMD and SyncNet-offset scores demonstrate that our method achieves more accurate lip-sync, while the F-LMD scores reflect that our method better restores facial expressions through intrinsic style. These results indicate that using intrinsic style can significantly enhance the quality of video generation.

To demonstrate the intrinsic style transfer capability of our method, we conducted expression transfer experiments in addition to video reconstruction experiments. This experiment can only be performed on the MEAD dataset, which contains multiple emotions for a single identity. Specifically, we used the neutral expression faces from the dataset as references and used videos with distinct expressions (such as happy, sad, and angry) to drive the generation. In such an experimental setup, methods that rely solely on audio for driving expressions struggle to effectively convey the corresponding emotions.
As shown in Table~\ref{tab:tab-cross}, \name~still leads in most metrics and maintains its superiority in this scenario. For the additionally calculated StyleSim, our method significantly outperforms other methods with intrinsic style, indicating that the transfer of intrinsic style can better control facial expressions and details.

\subsection{Qualitative Comparison}

In Figure~\ref{fig:stoa_cmp}, we present a visual comparison of our method with other methods, including comparisons of reconstruction results and intrinsic style transfer results.
In the reconstruction experiments, our method achieves accurate synchronization of head movements, lip shapes, and even eye blinking, while effectively preserving the identity of the speaker. In the intrinsic style transfer experiments, when there are significant differences in expressions between the reference face and the real video, our method effectively transfers the expressions and details of the face in the style reference video, while maintaining consistency in other conditions. In Figure~\ref{fig:demos}, we show the generated results of our method on different types of portraits. Our method successfully generates videos with rich expressions and natural movements on out-of-domain data, even non-human portraits, demonstrating strong robustness.

\subsection{Ablation Study}
\paragraph{Probabilistic Style Prior Learning.}
When training the style layer of the diffusion model in Sec.~\ref{sec:style-layer}, if we do not employ a probabilistic learning and instead use a deterministic style prior for training, it may lead to overfitting of the training results and lost identity information.
Figure~\ref{fig:cmp} shows the results of different intrinsic style acquisition methods. Using a deterministic intrinsic style causes the model to transfer (eye reflections/facial contours) from the style reference video to the new face, leading to issues with identity deviation. By employing the probabilistic learning and resampling method, the intrinsic style prior obtained by the model from the same training video varies each time, thereby preventing the transfer of incorrect content to the generated video.

\paragraph{Intrinsic Style Extractor with Audio Information.}
Table~\ref{tab:cluster_emo} provides a quantitative evaluation of the clustering strength of intrinsic style after incorporating audio features. We define clustering strength $d_{cls}$ as the ratio between inter-cluster distance $d_{inter}$ and intra-cluster distance $d_{intra}$:
\begin{equation}
    d_{cls} = \frac{d_{inter}}{d_{intra}}.
\end{equation}A larger value indicates a better clustering performance. We used different emotions of a single identity as categories to calculate the clustering strength of the intrinsic style obtained under three conditions: using only expression coefficients, using only audio features, and using both features together. The results indicate that expression coefficients play a crucial role in the extraction of intrinsic style and audio features can indeed serve as an auxiliary to enhance the clustering strength of intrinsic style, while using audio features alone is insufficient to obtain effective intrinsic style.

\paragraph{What Can We Learn from Style Prior?}
We use t-distributed Stochastic Neighbor Embedding (t-SNE)~\cite{van2008tsne} to project the intrinsic style priors into a two-dimensional space. Figure~\ref{fig:cluster} shows the intrinsic style priors of a speaker from the MEAD dataset. Each code is color-coded according to its corresponding emotion and intensity. The style priors with the same emotion first cluster together. Within each cluster, the style priors with the same intensity are closer to each other, and there are noticeable transitions between intrinsic style priors of different intensities. These observations indicate that our model can learn a continuous distribution of intrinsic styles. As shown in Figure ~\ref{fig:interpolation}, when performing linear interpolation between two intrinsic style priors extracted from the test set, the facial expressions and details in the generated video transition smoothly.
\section{Conclusion}
We propose \name~as the first talking head video generation method capable of achieving intrinsic style transfer. Through the design and training of the intrinsic style extractor, we obtain intrinsic style priors which can sufficiently represent the emotions and habits of the style reference videos. By sampling from the style prior and progressive training, we successfully transfer intrinsic styles to unseen faces. Experimental results show that \name~not only transfers intrinsic styles but also improves the overall quality of the generated videos, providing new insights for more advanced and comprehensive talking head video generation.

{
    \small
    \bibliographystyle{ieeenat_fullname}
    \bibliography{main}
}


\end{document}